\pdfoutput=1
\documentclass[sigconf,natbib=true]{acmart}
\AtBeginDocument{%
  }

\usepackage{multirow}
\usepackage{enumitem}

\definecolor{promptblock}{rgb}{0.0,0.35,0.75}
\definecolor{promptvar}{rgb}{0.0,0.55,0.15}
\definecolor{promptbg}{gray}{0.97}
\definecolor{promptframe}{gray}{0.45}
\newsavebox{\promptbox}
\newenvironment{promptblockbox}[1]
  {\setlength{\fboxrule}{0.6pt}\setlength{\fboxsep}{6pt}%
   \begin{lrbox}{\promptbox}
   \begin{minipage}{\dimexpr\columnwidth-2\fboxsep-2\fboxrule\relax}
   \footnotesize
   \begin{center}\small\bfseries #1\end{center}
   \vspace{-4pt}\rule{\linewidth}{0.4pt}\vspace{2pt}\par}
  {\end{minipage}\end{lrbox}%
   \fcolorbox{promptframe}{promptbg}{\usebox{\promptbox}}}
\usepackage{algorithm}
\usepackage{algorithmic}

\copyrightyear{2026}
\acmYear{2026}
\setcopyright{cc}
\setcctype{by}
\acmConference[CIKM '26]{Proceedings of the 35th ACM International Conference on Information and Knowledge Management}{November 07--11, 2026}{Rome, Italy}
\acmBooktitle{Proceedings of the 35th ACM International Conference on Information and Knowledge Management (CIKM '26), November 07--11, 2026, Rome, Italy}
\acmDOI{10.1145/3799682.3839887}
\acmISBN{979-8-4007-2539-5/2026/11}

\begin{document}

\title[RATED: Reliability-Aware AI Agent Tool Retrieval via Execution Dynamics]{RATED: Reliability-Aware AI Agent Tool Retrieval\\ via Execution Dynamics}

\author{Weiliang Chen}
\orcid{0000-0003-1978-7504}
\affiliation{%
  \department{Department of Computer Science}
  \institution{Texas Tech University}
  \city{Lubbock}
  \state{Texas}
  \country{USA}
}
\email{weilchen@ttu.edu}

\author{Song Liao}
\orcid{0000-0002-5264-7573}
\affiliation{%
  \department{Department of Computer Science}
  \institution{Texas Tech University}
  \city{Lubbock}
  \state{Texas}
  \country{USA}
}
\email{song.liao@ttu.edu}

\renewcommand{\shortauthors}{Weiliang Chen and Song Liao}

\begin{abstract}
As AI agents increasingly rely on external tools, accurate retrieval is essential for reliable behavior. Existing methods optimize semantic relevance but largely ignore tool availability during execution. To address this gap, we propose \textbf{RATED} (\textbf{R}eliability-\textbf{A}ware \textbf{T}ool retrieval via \textbf{E}xecution \textbf{D}ynamics), a retrieval framework that combines robust semantic matching with dynamic reliability adaptation. RATED employs a frozen dual-encoder with contrastive fine-tuning for semantic matching, and a lightweight contextual bandit module that re-ranks candidates based on real-time execution feedback. This design enables rapid adaptation to runtime tool failures without expensive model retraining. We introduce \textbf{EcoAgent}, a benchmark of 5,172 real-world tools and over 210,000 query--tool pairs for offline retrieval and dynamic reliability evaluation. Experimental results demonstrate that RATED achieves a 52.2\% relative MRR improvement over the ToolLLM API retriever on real-world queries, and reduces fault selection rate from 86.2\% to 2.9\% under severe API outage scenarios with a 30\% failure rate, while incurring negligible latency overhead of less than 1\,ms over encode-only inference. Code and dataset are available at \url{https://github.com/Elliotes/EcoAgent}.
\end{abstract}



\maketitle
\section{Introduction}
\label{sec:introduction}
Large Language Models (LLMs) have demonstrated impressive capabilities in natural language processing but are inherently unable to execute real-world actions~\cite{10.1145/3735632,10.1145/3744746}. To overcome this limitation, recent systems integrate LLMs with external tools and APIs~\cite{mialon2023augmentedlanguagemodelssurvey,schick2023toolformerlanguagemodelsteach}, yielding autonomous AI agents that fulfill actionable requests, such as querying live databases or managing enterprise workflows~\cite{10748304}. The fundamental bottleneck for these systems has shifted from text generation to accurately identifying and retrieving the correct entry from vast, growing tool repositories~\cite{patil2023gorillalargelanguagemodel}.
\begin{figure}[!t]
  \centering
  \includegraphics[width=\columnwidth]{"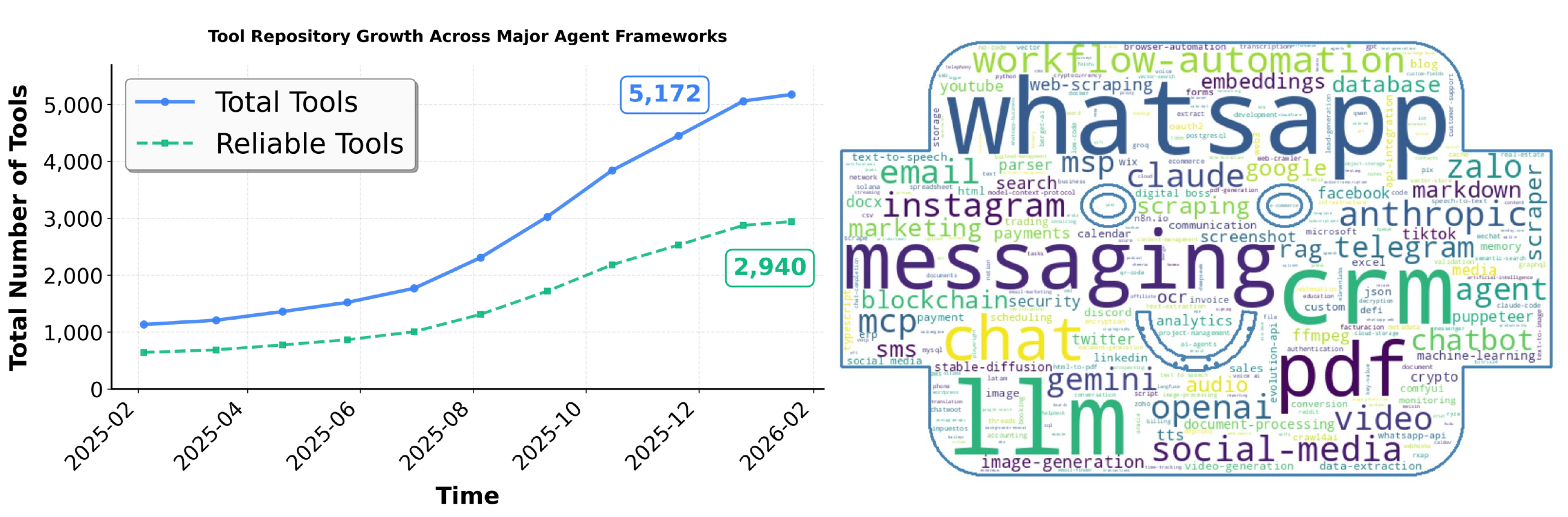"}
  \Description{Left panel shows tool repository growth timeline from February 2025 to January 2026, with total tools (blue line) and reliable tools (orange line). Right panel shows tag word cloud visualization.}
  \vspace{-0.8cm}
  \caption{Tool repository growth and reliability gap on the left, and skewed tag distribution on the right.}
  \label{fig:tool_challenges}
  \vspace{-0.3cm}
\end{figure}
To address the tool retrieval challenge, existing research has focused on optimizing semantic relevance for Large Language Models (LLMs). Foundational works like ToolLLM~\cite{qin2023toolllm} and ToolExpNet~\cite{zhang2025toolexpnet} treat tool retrieval as a static ranking task. They use dense retrieval or experience networks to map queries to tool documentation based entirely on semantic relevance. While approaches such as TRFeedback~\cite{xu2024trfeedback} and ToolRL~\cite{qian2025toolrl} incorporate execution feedback, they still aim to optimize how the LLM plans and invokes tools in a static environment.
However, real AI agent tool ecosystems are far from static and highly unreliable. We crawl tool catalogs from LangChain~\cite{langchain2023}, CrewAI~\cite{crewai2024}, LlamaIndex~\cite{llamaindex2024}, and n8n~\cite{n8n2024} to reveal the severity of this issue. As Figure~\ref{fig:tool_challenges} shows, despite rapid growth, over 50\% of these tools exhibit critical flaws like broken links and missing authentication. Many tools are functionally similar, so retrieval cannot rely on semantics alone to identify which ones will execute successfully.
To overcome this challenge, the system must adapt dynamically during execution. When a tool fails at runtime, the system must promptly adjust candidate ordering between requests. ToolRL cannot achieve this prompt adjustment because it uses Group Relative Policy Optimization (GRPO)~\cite{GRPO} to update the LLM from batched trajectories and still targets the planner. Contextual bandits~\cite{li2010linucb} effectively address this problem. They condition each decision on the current query and update from each individual execution outcome. Moreover, they observe feedback only for the selected tool and make immediate adjustments without retraining the semantic retriever.

In this paper, we propose \textbf{RATED} (\textbf{R}eliability-\textbf{A}ware \textbf{T}ool retrieval via \textbf{E}xecution \textbf{D}ynamics), a retrieval framework that dynamically adapts to runtime tool failures without human intervention. First, we introduce \textbf{EcoAgent}, a benchmark of 5,172 real-world agent tools and over 210,000 query--tool pairs. Second, RATED employs a dual-encoder fine-tuned with a contrastive objective for semantic matching, together with a lightweight contextual bandit that re-ranks the top-$K$ candidates using a prior based on metadata and execution feedback, enabling the system to avoid failing tools. Experiments show that RATED alone keeps fault selection below 3\% and recovers at both failure rates. Under a 30\% failure injection protocol, RATED reduces fault selection rate to 2.9\%, compared with 86.2\% for the ToolLLM API retriever, while adding only 0.172\,ms latency per step over encode-only inference.

\section{The EcoAgent Benchmark}
\label{sec:dataset}

We construct EcoAgent, a benchmark comprising two complementary datasets. EcoAgent complements existing tool-use datasets with metadata reliability scores for every tool and with an evaluation protocol for injected execution failures. EcoAgent-Syn supports robust training with synthetic pairs, and EcoAgent-Real supports rigorous evaluation using real-world queries. We aggregate 5,172 real-world tool definitions from four open-source AI agent platforms, namely LangChain, CrewAI, LlamaIndex, and n8n. Metadata are normalized into a shared schema with functional descriptions, names, endpoints, and authentication fields, which supports retrieval text construction and reliability scoring. To operationalize tool quality, each tool receives an automated metadata reliability score between 0 and 1 from three weighted signals, namely name--description consistency, source credibility, and documentation quality, which Section~\ref{sec:online_adaptation} defines. The benchmark does not rely on manual per-tool ratings. Tools with scores of at least 0.8 are marked as reliable, identifying 2,940 tools, or 56.8\% of the repository. The encoder is trained only on EcoAgent-Syn, while retrieval and dynamic experiment results are validated on the EcoAgent-Real test split, with hyperparameters selected on the validation split.

We build EcoAgent-Syn with Gemma-3-1b~\cite{geminiteam2025geminifamilyhighlycapable}. For each tool, the model receives the tool name, functional description, and metadata as context and generates 10--20 task-oriented queries. Queries are restricted to 5--120 words, and near-duplicates with Jaccard similarity above 0.7 are removed to preserve intent diversity. Figure~\ref{fig:prompt} shows the prompt template, which we also release with the dataset. To improve robustness against spurious correlations, each query is paired with hard negatives from SBERT~\cite{reimers2019sentence} retrieval and random negatives. EcoAgent-Syn links 69,501 unique synthetic queries to 5,172 ground-truth tools, with 70,122 positive pairs and 140,244 negative pairs. For evaluation under realistic language distributions, we build EcoAgent-Real from ToolBench~\cite{qin2023toolllm}. Because the original dataset lacks ground-truth mappings to our tool repository, we implement a semi-automated labeling pipeline. GTE-Large~\cite{li2023generaltextembeddingsmultistage} retrieves the top-5 candidates per query, discards pairs with cosine similarity below 0.7, and Gemma-3-1b verifies the remainder, retaining only pairs with functional match scores of at least 0.85. Human experts review a stratified sample of 200 queries and agree with the LLM verifier on 94\% of cases. After filtering and deduplication, EcoAgent-Real contains 6,379 unique queries and 9,946 verified query--tool pairs.

\begin{figure}[!t]
\centering
\begin{promptblockbox}{Prompt for Query Synthesis}
\textcolor{promptblock}{\textbf{[Task]}}~~You are generating diverse user queries
for an AI agent tool recommendation system.
\smallskip

\textcolor{promptblock}{\textbf{[Inputs]}}~~
\textcolor{promptvar}{\texttt{\{tool\_name\}}} the tool name~~\textbar~~
\textcolor{promptvar}{\texttt{\{tool\_description\}}} its functional description~~\textbar~~
\textcolor{promptvar}{\texttt{\{num\_queries\}}} how many queries to write
\smallskip

\textcolor{promptblock}{\textbf{[Requirements]}}
\begin{enumerate}[nosep,leftmargin=1.6em,topsep=2pt]
  \item Each query should be 5-120 words.
  \item Vary the intent types: direct action requests, troubleshooting questions, how-to questions.
  \item Use natural, conversational language.
  \item DO NOT copy phrases verbatim from the tool description.
  \item Make queries task-oriented and practical.
\end{enumerate}
\smallskip

\textcolor{promptblock}{\textbf{[Output]}}~~One query per line, no numbering or bullets.
\end{promptblockbox}
\Description{A boxed prompt template with four labelled parts. The task line asks the model to generate diverse user queries for an AI agent tool recommendation system. The inputs are the tool name, the tool description, and the number of queries to write. Five numbered requirements ask for queries of 5 to 120 words, varied intent types, natural language, no verbatim copying from the tool description, and task-oriented phrasing. The output line asks for one query per line without numbering or bullets.}
  \vspace{-0.3cm}
\caption{Prompt template that generates the EcoAgent-Syn queries.}
\label{fig:prompt}
  \vspace{-0.3cm}
\end{figure}

\begin{figure*}[t!]
  \centering
  \includegraphics[width=0.9\textwidth]{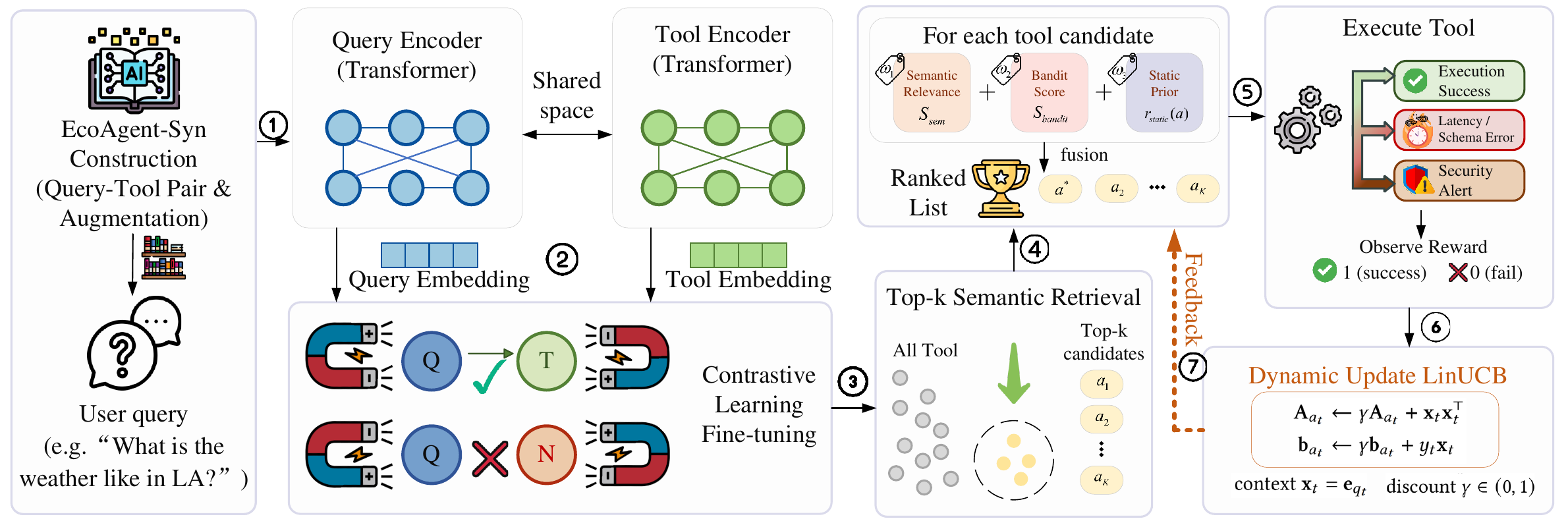}
  \Description{RATED architecture with DRAGON dual-encoder retrieval, bandit re-ranking, and execution feedback.}
  \vspace{-0.2cm}
  \caption{RATED architecture with DRAGON retrieval, bandit re-ranking, and execution feedback.}
  \label{fig:framework_overview}
\end{figure*}

\section{Methodology}
\label{sec:methodology}

\subsection{Problem Setup}
\label{sec:problem_and_framework}
We formulate agent tool retrieval as sequential decision making over a shared repository. Let $\mathcal{U} = \{a_1, \dots, a_M\}$ denote the tool pool, where $M{=}5{,}172$ in EcoAgent. At step $t$, the agent issues query $q_t$. The retriever returns a candidate set $\mathcal{A}_t \subset \mathcal{U}$ with $|\mathcal{A}_t| = K$, and the policy selects one tool $a_t \in \mathcal{A}_t$ for execution. The objective is to maximize the probability of successful execution. Unlike static retrieval with a fixed mapping $f(q, a) \rightarrow \{0,1\}$, production agents must cope with tools that fail at runtime, where tool utility depends on semantic relevance $S_{\text{sem}}(q_t, a)$ and time-varying reliability. After execution, only the chosen tool yields binary feedback $y_t \in \{0, 1\}$, yielding partial feedback that we address with contextual bandits~\cite{10.1007/s00453-003-1038-1}. Figure~\ref{fig:framework_overview} summarizes the RATED pipeline.

\subsection{Semantic Perception Module}
\label{sec:semantic_perception}

The semantic perception module implements the first stage of RATED. Given $q_t$, it scores all tools in $\mathcal{U}$ by $S_{\text{sem}}$ and returns the top-$K$ candidate set $\mathcal{A}_t$ defined in Section~\ref{sec:problem_and_framework}. The semantic encoder is based on DRAGON~\cite{lin-etal-2023-dragon}, fine-tuned on EcoAgent-Syn and frozen at deployment. New tools can be added by encoding their documentation with $f_T$ without retraining the encoders.

\subsubsection{DRAGON Dual-Encoder Architecture}
DRAGON provides separate query encoder $f_Q$ and tool encoder $f_T$, which suits settings where user queries and tool documentation follow different lexical patterns. For query $q$ and tool $a$, we take CLS representations and L2-normalize them. The semantic relevance score is the dot product of the normalized vectors.
\begin{equation}
\mathbf{e}_q = \mathrm{norm}(f_Q(q)), \quad \mathbf{e}_a = \mathrm{norm}(f_T(a))
\end{equation}
\begin{equation}
S_{\text{sem}}(q, a) = \mathbf{e}_q^\top \mathbf{e}_a
\end{equation}

\subsubsection{Contrastive Fine-tuning on EcoAgent-Syn}
Although DRAGON is effective for open-domain retrieval, agent tool descriptions use specialized API terminology. We fine-tune $f_Q$ and $f_T$ on EcoAgent-Syn with an in-batch InfoNCE objective~\cite{oord2019representationlearningcontrastivepredictive}. For a batch of $B$ positive pairs $\{(q_i, a_i^+)\}_{i=1}^B$, we optimize
\begin{equation}
\mathcal{L}_{\text{InfoNCE}} = -\frac{1}{B}\sum_{i=1}^{B} \log \frac{\exp(\mathbf{e}_{q_i}^\top \mathbf{e}_{a_i^+} / \tau)}{\sum_{j=1}^{B} \exp(\mathbf{e}_{q_i}^\top \mathbf{e}_{a_j^+} / \tau)}
\label{eq:infonce_loss}
\end{equation}
where $\tau$ is a temperature hyperparameter. Hard negatives from EcoAgent-Syn construction in Section~\ref{sec:dataset} are used in training. In-batch negatives provide additional contrast during optimization.

\subsection{Reliability-Aware Re-ranking}
\label{sec:online_adaptation}

The second stage re-ranks $\mathcal{A}_t$ from Section~\ref{sec:semantic_perception} using operational signals while the semantic encoders remain frozen. At step $t$, the context is the query embedding $\mathbf{x}_t = \mathbf{e}_{q_t} \in \mathbb{R}^d$, the policy selects $a_t \in \mathcal{A}_t$, and observes reward $y_t \in \{0,1\}$ only after execution, where $y_t{=}1$ denotes success and $y_t{=}0$ denotes errors or timeouts.

Tools with little execution history are initialized from catalog metadata. We define a static prior $r_{\text{static}}(a) \in [0,1]$ from three signals used in benchmark construction in Section~\ref{sec:dataset}, namely name--description consistency $r_{\text{cons}}(a)$, source credibility $w_{\text{src}}(a)$, and documentation quality $r_{\text{qual}}(a)$.
\begin{equation}
r_{\text{cons}}(a) = \cos(f_{\text{Enc}}(a_{\text{name}}), f_{\text{Enc}}(a_{\text{desc}}))
\end{equation}
\begin{equation}
r_{\text{static}}(a) = \lambda_1 r_{\text{cons}}(a) + \lambda_2 w_{\text{src}}(a) + \lambda_3 r_{\text{qual}}(a)
\label{eq:static_prior}
\end{equation}
where $\sum_i \lambda_i = 1$. These signals implement the catalog reliability score in Section~\ref{sec:dataset}, and all three read catalog fields only, so no manual rating is needed. For $r_{\text{cons}}$, $f_{\text{Enc}}$ is the SBERT encoder of Section~\ref{sec:dataset}, which embeds the tool name and its description, and we take the cosine of the two normalized vectors. For $w_{\text{src}}$, each platform carries a fixed weight, namely 1.0 for an official listing, 0.9 for LangChain, CrewAI, and LlamaIndex, 0.8 for n8n, and 0.6 otherwise. For $r_{\text{qual}}$, we standardize the description length and the tag-string length against their medians and pass $0.6\,z_{\text{desc}} + 0.4\,z_{\text{tag}}$ through a logistic function. The 0.8 threshold is used for benchmark statistics, while $r_{\text{static}}$ initializes the bandit. Each arm bias is initialized as $\mathbf{b}_a \leftarrow r_{\text{static}}(a)\cdot \mathbf{x}_0$, which covers the cold-start case, where a new or rarely used tool carries no execution history and is therefore scored from its catalog metadata.
\begin{algorithm}[htb]
\caption{Reliability-aware adaptation via discounted LinUCB}
\label{alg:discounted_linucb}
\begin{algorithmic}[1]
\REQUIRE query stream $\{q_t\}_{t=1}^{T}$, frozen encoders $f_Q$ and $f_T$, discount $\gamma$, exploration $\alpha$, fusion weights $w_1, w_2, w_3$
\ENSURE tool $a_t$ executed at each step, with $(\mathbf{A}_a, \mathbf{b}_a)$ updated online
\STATE $\mathbf{e}_a \leftarrow \mathrm{norm}(f_T(a))$ for all $a \in \mathcal{U}$ \COMMENT{offline, reused at every step}
\STATE $\mathbf{A}_a \leftarrow \mathbf{I}_d$ and $\mathbf{b}_a \leftarrow r_{\text{static}}(a)\,\mathbf{x}_0$ \COMMENT{cold start from metadata}
\FOR{$t = 1$ \TO $T$}
  \STATE $\mathbf{x}_t \leftarrow \mathrm{norm}(f_Q(q_t))$
  \STATE $\mathcal{A}_t \leftarrow$ top-$K$ tools by $S_{\text{sem}}(q_t, a) = \mathbf{x}_t^\top \mathbf{e}_a$
  \FOR{each $a \in \mathcal{A}_t$}
    \STATE $\hat{\boldsymbol{\theta}}_a \leftarrow \mathbf{A}_a^{-1} \mathbf{b}_a$
    \STATE $S_{\text{bandit}} \leftarrow \hat{\boldsymbol{\theta}}_a^\top \mathbf{x}_t + \alpha \sqrt{\mathbf{x}_t^\top \mathbf{A}_a^{-1} \mathbf{x}_t}$
    \STATE $S_{\text{final}}(a) \leftarrow w_1 S_{\text{sem}} + w_2 S_{\text{bandit}} + w_3 r_{\text{static}}(a)$
  \ENDFOR
  \STATE $a_t \leftarrow \arg\max_{a \in \mathcal{A}_t} S_{\text{final}}(a)$
  \STATE execute $a_t$ and observe $y_t \in \{0,1\}$
  \STATE $\mathbf{A}_{a_t} \leftarrow \gamma \mathbf{A}_{a_t} + \mathbf{x}_t \mathbf{x}_t^\top$ \COMMENT{kept diagonal, so scoring is $O(d)$}
  \STATE $\mathbf{b}_{a_t} \leftarrow \gamma \mathbf{b}_{a_t} + y_t \mathbf{x}_t$
\ENDFOR
\end{algorithmic}
\end{algorithm}

At runtime, the system must adapt to dynamic failures such as API outages. Standard LinUCB~\cite{li2010linucb} weights all past observations equally and therefore adapts slowly when arm quality shifts over time. Alternatively, a global multi-armed bandit (Global MAB)~\cite{GlobalMAB} tracks shared or context-free arm statistics. It is cheaper to maintain but cannot model payoffs that depend on the current query context. To overcome these limitations, RATED adopts discounted LinUCB~\cite{10.5555/2050345.2050365} with a separate linear model per tool. This choice addresses non-stationarity by down-weighting older rewards before each update, ensuring that recent failures change tool scores more quickly. Each tool $a$ maintains $\mathbf{A}_a \in \mathbb{R}^{d \times d}$ and $\mathbf{b}_a \in \mathbb{R}^d$. The reliability score combines exploitation and exploration.
\begin{equation}
S_{\text{bandit}}(\mathbf{x}_t, a) = \hat{\boldsymbol{\theta}}_a^\top \mathbf{x}_t + \alpha \sqrt{\mathbf{x}_t^\top \mathbf{A}_a^{-1} \mathbf{x}_t}
\label{eq:bandit_score}
\end{equation}
where $\hat{\boldsymbol{\theta}}_a = \mathbf{A}_a^{-1} \mathbf{b}_a$. After executing $a_t$ and observing $y_t$, we update with discount $\gamma \in (0, 1)$.
\begin{align}
\mathbf{A}_{a_t} &\leftarrow \gamma \mathbf{A}_{a_t} + \mathbf{x}_t \mathbf{x}_t^\top \\
\mathbf{b}_{a_t} &\leftarrow \gamma \mathbf{b}_{a_t} + y_t \mathbf{x}_t
\end{align}
Discounting emphasizes recent feedback and encourages timely switches to alternatives.

\subsection{Hybrid Fusion for Tool Ranking}
\label{sec:hybrid_scoring}
RATED fuses semantic, bandit, and metadata signals into one score over $\mathcal{A}_t$. For query $q_t$ and candidate $a \in \mathcal{A}_t$, we define
\begin{equation}
S_{\text{final}}(q_t, a) = w_1 S_{\text{sem}}(q_t, a) + w_2 S_{\text{bandit}}(\mathbf{x}_t, a) + w_3 r_{\text{static}}(a)
\label{eq:hybrid_score}
\end{equation}
where $w_1 + w_2 + w_3 = 1$. Weights are tuned on the validation split and fixed for testing in Section~\ref{sec:experiments}. At each step, the system retrieves top-$K$ tools by $S_{\text{sem}}$, ranks $\mathcal{A}_t$ by $S_{\text{final}}$, executes the highest-scoring tool, and updates bandit parameters from $y_t$. Algorithm~\ref{alg:discounted_linucb} summarizes one step of this loop.

\section{Experiments}
\label{sec:experiments}

\subsection{Experimental Setup}

All experiments run on a single NVIDIA GeForce RTX 5070 GPU (12\,GB VRAM). DRAGON is fine-tuned on EcoAgent-Syn for one epoch with batch size 8, learning rate $1 \times 10^{-5}$, 1{,}000 warmup steps, $\tau = 0.05$, mixed precision, and maximum sequence length 256. At inference, the system retrieves top-$K{=}10$ tools by $S_{\text{sem}}$ and re-ranks them with Eq.~\ref{eq:hybrid_score}. The bandit uses $\alpha = 0.1$, $\gamma = 0.95$, and 2{,}000 warm-up pairs, with diagonal covariance updates for $O(d)$ scoring at each step. Fusion weights tuned on the EcoAgent-Real validation split are $(w_1, w_2, w_3) = (0.75, 0.15, 0.10)$. Static weights $(\lambda_1, \lambda_2, \lambda_3)$ in Eq.~\ref{eq:static_prior} are $(0.5, 0.3, 0.2)$, and biases are initialized with $\mathbf{x}_0 = \mathbf{1}/\sqrt{d}$. 
\begin{table}[!t]
  \centering
  \caption{Retrieval performance on EcoAgent-Syn and EcoAgent-Real.}
  \label{tab:results}
  \footnotesize
  \setlength{\tabcolsep}{3pt}
  \renewcommand{\arraystretch}{1.1}
  \resizebox{\columnwidth}{!}{%
  \begin{tabular}{lcccccc}
    \toprule
    \multirow{2}{*}{Method} & \multicolumn{3}{c}{EcoAgent-Syn} & \multicolumn{3}{c}{EcoAgent-Real} \\
    \cmidrule(lr){2-4} \cmidrule(lr){5-7}
     & MRR & Recall@1 & Recall@5 & MRR & Recall@1 & Recall@5 \\
    \midrule
    TF-IDF & 0.680 & 0.607 & 0.769 & 0.274 & 0.116 & 0.326 \\
    BM25 & 0.235 & 0.178 & 0.300 & 0.169 & 0.078 & 0.194 \\
    OpenAI Ada-002 & 0.790 & 0.736 & 0.857 & 0.670 & 0.430 & 0.750 \\
    BGE-M3 & 0.765 & 0.702 & 0.840 & 0.511 & 0.289 & 0.579 \\
    GTE-Large & 0.772 & 0.713 & 0.842 & 0.515 & 0.294 & 0.581 \\
    E5-Large-v2 & 0.746 & 0.674 & 0.830 & 0.597 & 0.466 & 0.756 \\
    DRAGON (pretrained) & 0.751 & 0.690 & 0.823 & 0.518 & 0.407 & 0.654 \\
    SPLADE-v3 & 0.754 & 0.689 & 0.834 & 0.626 & 0.491 & 0.789 \\
    BGE-Reranker & 0.596 & 0.483 & 0.735 & 0.642 & 0.491 & 0.840 \\
    ToolLLM API Retriever & 0.724 & 0.666 & 0.795 & 0.406 & 0.222 & 0.409 \\
    \midrule
    RATED (w/o Contrastive) & 0.751 & 0.690 & 0.823 & 0.518 & 0.407 & 0.654 \\
    RATED (Full) & \textbf{0.845} & \textbf{0.804} & \textbf{0.893} & 0.618 & 0.516 & 0.740 \\
    \bottomrule
  \end{tabular}%
  }
\end{table}
\subsection{Evaluation Metrics}
\label{sec:metrics}

Offline retrieval is measured with MRR, Recall@1, and Recall@5 on the EcoAgent-Real test split. EcoAgent-Syn uses the same metrics for in-domain comparison.
Dynamic reliability is evaluated over $T{=}10{,}000$ sequential selections. At step $t$, the policy selects $a_t$ for query $q_t$ with ground truth $a^*_t$. The indicator $\text{Status}(a,t)\in\{0,1\}$ denotes whether tool $a$ is available under simulated failure injection. To simulate a persistent outage, the simulator ranks the $N$ distinct ground-truth tools of the evaluation stream by frequency, skips the top 10\%, and marks the next $\rho N$ tools as failing, where $\rho$ is the failure rate. The failing set is fixed before the run, and its tools stay unavailable from failure onset to the end. Metrics are averaged over the window after failure onset $\mathcal{T}_{\text{post}}=\{2{,}001,\ldots,5{,}000\}$, which contains $T_{\text{post}}{=}3{,}000$ steps.
\begin{equation}
\text{Post-Fail SR} = \frac{1}{T_{\text{post}}} \sum_{t \in \mathcal{T}_{\text{post}}} \mathbb{I}(a_t = a^*_t) \cdot \text{Status}(a_t, t),
\label{eq:post_fail_sr}
\end{equation}
\begin{equation}
\text{FSR} = \frac{1}{T_{\text{post}}} \sum_{t \in \mathcal{T}_{\text{post}}} \mathbb{I}(\text{Status}(a_t, t) = 0),
\label{eq:fault_rate}
\end{equation}
where FSR is the fault selection rate. Adaptation steps (AS) measure recovery speed using
\begin{equation}
t_{\text{adapt}} = \min \{t : \text{SR}_{\text{window}}(t) \geq \theta \times \text{SR}_{\text{peak}}\},
\label{eq:adaptation_steps}
\end{equation}
computed from failure onset with a sliding window of $W{=}200$ steps, peak success rate before failure $\text{SR}_{\text{peak}}$, and $\theta{=}0.9$ at 20\% failure rate or $\theta{=}0.8$ at 30\%. Static rankers without online feedback never reach the recovery target and are marked \textit{Failed}.

\subsection{Baselines}

Retrieval baselines include lexical matchers TF-IDF~\cite{sparck1972statistical} and BM25~\cite{10.5555/188490.188561}, dense encoders OpenAI Ada-002~\cite{openai2022embeddings}, BGE-M3~\cite{bge-m3}, GTE-Large~\cite{li2023generaltextembeddingsmultistage}, and E5-Large-v2~\cite{wang2022e5}, and neural IR models DRAGON~\cite{lin-etal-2023-dragon}, which is not fine-tuned on EcoAgent-Syn, and SPLADE-v3~\cite{lassance2024spladev3}. We also include a BGE-Reranker-v2-m3~\cite{bge-m3} cross-encoder that re-ranks DRAGON top-50 candidates, and the ToolLLM API retriever~\cite{qin2023toolllm}. For dynamic evaluation, two reactive baselines share the DRAGON ranker: Health Check Blacklist permanently excludes any tool after one failed execution, and Retry ($k{=}3$) tries up to three candidates within one step and keeps no memory across steps. A batch GRPO proxy similar to ToolRL~\cite{GRPO} updates weights periodically rather than online. RATED uses DRAGON fine-tuned on EcoAgent-Syn as the backbone and is evaluated as RATED (Full), RATED (w/o Contrastive), RATED (w/o Feedback), and RATED (w/ Global MAB).

\begin{table}[t!]
  \centering
  \caption{Dynamic reliability evaluation under injected tool failures.}
  \label{tab:dynamic_reliability}
  \footnotesize
  \setlength{\tabcolsep}{3pt}
  \renewcommand{\arraystretch}{1.1}
  \resizebox{\columnwidth}{!}{%
  \begin{tabular}{lcccccc}
    \toprule
    \multirow{2}{*}{Method} & \multicolumn{3}{c}{Failure Rate 20\%} & \multicolumn{3}{c}{Failure Rate 30\%} \\
    \cmidrule(lr){2-4} \cmidrule(lr){5-7}
    & Post-Fail SR (\%) $\uparrow$ & FSR (\%) $\downarrow$ & AS $\downarrow$ & Post-Fail SR (\%) $\uparrow$ & FSR (\%) $\downarrow$ & AS $\downarrow$ \\
    \midrule
    BM25 & 18.8 & 20.0 & \textit{Failed} & 10.2 & 30.0 & \textit{Failed} \\
    ToolLLM API Retriever & 17.4 & 79.6 & \textit{Failed} & 10.5 & 86.2 & \textit{Failed} \\
    OpenAI Ada-002 & 19.4 & 80.4 & \textit{Failed} & 11.1 & 88.8 & \textit{Failed} \\
    DRAGON (pretrained) & 78.1 & 18.1 & \textit{Failed} & 75.7 & 21.2 & \textit{Failed} \\
    SPLADE-v3 & 16.0 & 7.7 & \textit{Failed} & 15.8 & 9.1 & \textit{Failed} \\
    BGE-Reranker & 83.2 & 17.5 & \textit{Failed} & 81.0 & 20.5 & \textit{Failed} \\
    \midrule
    Health Check Blacklist & 82.6 & 1.4 & 2{,}620 & 80.2 & 1.9 & \textit{Failed} \\
    Retry ($k{=}3$) & 84.8 & 16.4 & 70 & 82.9 & 19.1 & \textit{Failed} \\
    \midrule
    ToolRL-like Batch GRPO & 80.3 & 5.4 & 2{,}620 & 79.5 & 6.1 & 1 \\
    RATED (w/o Feedback) & 81.1 & 16.5 & \textit{Failed} & 78.7 & 19.4 & \textit{Failed} \\
    RATED (w/ Global MAB) & 78.1 & 10.1 & 3{,}852 & 74.3 & 13.2 & 2{,}718 \\
    \textbf{RATED (Full)} & \textbf{83.4} & \textbf{2.5} & \textbf{1{,}603} & \textbf{82.1} & \textbf{2.9} & \textbf{0} \\
    \bottomrule
  \end{tabular}%
  }
\end{table}
\subsection{Main Results}
\label{sec:dynamic_reliability}

Table~\ref{tab:results} reports offline retrieval on EcoAgent-Syn and the held-out EcoAgent-Real test split. RATED (Full) ranks first on Syn across all metrics, and RATED (w/o Contrastive) matches pretrained DRAGON on both benchmarks, which isolates the gain from contrastive fine-tuning. On EcoAgent-Real, RATED (Full) improves MRR by 52.2\% over the ToolLLM API retriever but stays below Ada-002 and SPLADE-v3. High offline ranking accuracy on real queries therefore does not guarantee robust tool selection when tools fail at runtime.

Table~\ref{tab:dynamic_reliability} reports Post-Fail SR, FSR, and adaptation steps under 20\% and 30\% failure injection. Static retrievers, including ToolLLM and Ada-002, frequently select unavailable tools and never recover within the evaluation horizon. DRAGON (pretrained) and BGE-Reranker reduce FSR but remain unable to adapt online. Relative to RATED (w/o Feedback), RATED (Full) lowers FSR to 2.5\% and 2.9\% while it maintains Post-Fail SR above 82\%, which is a $7.3\times$ reduction over DRAGON (pretrained) and a $30\times$ reduction over ToolLLM. Batch GRPO updates similar to ToolRL retain higher FSR, and health check blacklisting attains low FSR but recovers slowly. RATED (w/ Global MAB) stays inferior to the full model. These results support decoupling semantic matching from reliability estimation updated online from execution feedback.

\subsection{Hyperparameter Sensitivity}
\label{sec:sensitivity}

\begin{figure}[!t]
  \centering
  \includegraphics[width=\columnwidth]{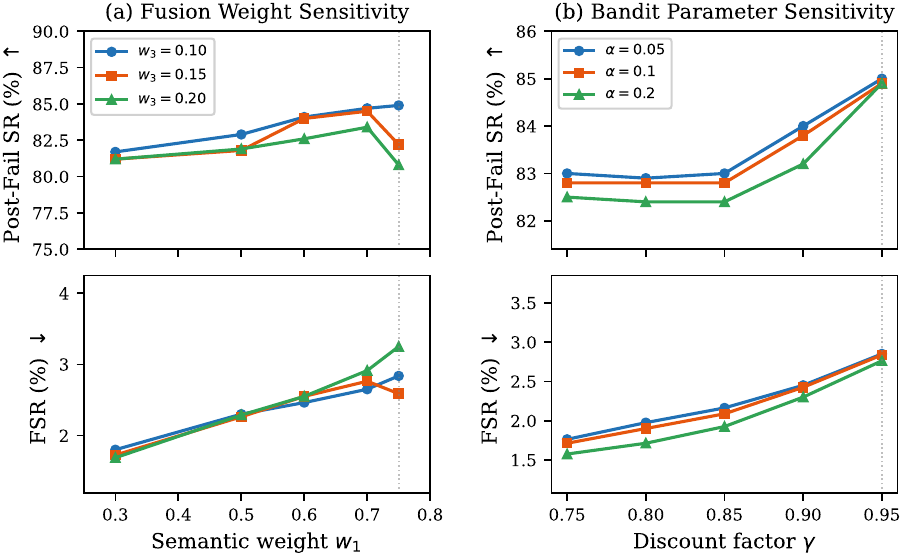}
  \Description{Two-panel sensitivity analysis under 20 percent failure injection. Left panel plots fault selection rate versus semantic weight w1 for three static-prior weights w3. Right panel plots fault selection rate versus discount factor gamma for three exploration coefficients alpha. Vertical dotted lines mark default hyperparameters.}
  \caption{Hyperparameter sensitivity at 20\% failure injection.}
  \label{fig:sensitivity}
\end{figure}

Figure~\ref{fig:sensitivity} reports Post-Fail SR and FSR under 20\% failure injection for two grid sweeps over fusion and bandit hyperparameters.
Panel~(a) varies $w_1$ and $w_3$ with bandit settings fixed. Post-Fail SR rises with $w_1$ and peaks at $w_1 = 0.75$ with $w_3 = 0.10$, whereas FSR grows slowly over the same range and stays in the low single-digit range throughout. We therefore select $w_1 = 0.75$ and $w_3 = 0.10$ on the validation split, which trades about one point of FSR for the highest Post-Fail SR.
Panel~(b) varies $\gamma$ and $\alpha$ with fusion weights fixed at the validated setting. Post-Fail SR is highest at $\gamma = 0.95$, where the three $\alpha$ curves nearly coincide, and FSR again grows slowly with $\gamma$. We therefore set $\gamma = 0.95$ and $\alpha = 0.1$ on the validation split.

\subsection{Efficiency Analysis}
\label{sec:efficiency}

We profile end-to-end inference at batch size 1 on the same hardware as Section~\ref{sec:experiments}. Table~\ref{tab:efficiency_detailed} reports average latency, P95 latency, and incremental overhead relative to DRAGON encoding only.
The encode-only baseline averages 3.223\,ms. RATED (w/o Feedback) applies only the static reliability prior in Eq.~\ref{eq:static_prior} and adds 0.025\,ms. Contextual bandit re-ranking incurs modest additional cost. RATED (w/ Global MAB) adds 0.108\,ms, and RATED (Full) adds 0.172\,ms, reaching 3.395\,ms average and 4.391\,ms P95 latency. These overheads are far below batch GRPO updates similar to ToolRL at 17.839\,ms average, which require periodic policy synchronization. Therefore, RATED is fast enough for interactive agent loops and still reduces fault selection.
\vspace{-0.2cm}
\begin{table}[h]
  \centering
  \caption{Inference latency at batch size 1 (ms).}
  \label{tab:efficiency_detailed}
  \footnotesize
  \setlength{\tabcolsep}{3pt}
  \renewcommand{\arraystretch}{1.1}
  \begin{tabular}{lccc}
    \toprule
    Method & Avg.\ Latency & P95 Latency & $\Delta$ Overhead \\
    \midrule
    DRAGON encode only & 3.223 & 4.269 & -- \\
    ToolRL-like Batch GRPO & 17.839 & 28.921 & +14.616 \\
    \midrule
    RATED (w/o Feedback) & 3.248 & 4.274 & +0.025 \\
    RATED (w/ Global MAB) & 3.331 & 4.351 & +0.108 \\
    \textbf{RATED (Full)} & \textbf{3.395} & \textbf{4.391} & \textbf{+0.172} \\
    \bottomrule
  \end{tabular}
\end{table}

\section{Conclusion}
\label{sec:conclusion}

We introduced RATED, a reliability-aware tool retrieval framework that combines DRAGON encoders fine-tuned with contrastive learning and contextual bandit re-ranking from execution feedback. On EcoAgent, RATED achieves strong in-domain retrieval and remains competitive on real-world queries. Under injected tool failures, it reduces fault selection to 2.5--2.9\% FSR, $7.3\times$ lower than static DRAGON, with only 0.172\,ms added latency over encode-only inference. These results highlight the critical importance of incorporating execution feedback into retrieval systems for autonomous agents operating in dynamic environments, and demonstrate that RATED provides an efficient solution for reliable tool discovery in agent systems. Evaluating this cold-start path on newly added tools and validating RATED in live agent runtimes remain future work.

\begin{acks}
This work is supported by the National Science Foundation (NSF) under award number 2623261.
\end{acks}

\section*{GenAI Usage Disclosure}

We use Gemma-3-1b to generate the synthetic queries of EcoAgent-Syn and to verify candidate query--tool pairs during EcoAgent-Real labeling, and we release the prompt templates with the dataset. We used OpenAI's ChatGPT to improve the clarity and grammar of sentences during the writing process.

\bibliographystyle{ACM-Reference-Format}
\balance
\bibliography{sample-base}

\end{document}